\documentclass[11pt]{article}
\usepackage{acl}

\usepackage{times}
\usepackage{latexsym}
\usepackage[T1]{fontenc}
\usepackage[utf8]{inputenc}
\usepackage{microtype}
\usepackage{inconsolata}
\usepackage{amsmath,amssymb,mathtools}
\usepackage{booktabs,multirow,tabularx,array}
\usepackage{graphicx}
\usepackage{algorithm}
\usepackage{algorithmic}
\usepackage{enumitem}
\usepackage{xspace}
\usepackage{xcolor}
\usepackage{tikz}
\usepackage{pgfplots}
\usepgfplotslibrary{groupplots}
\usetikzlibrary{arrows.meta,positioning,calc,fit,backgrounds}
\pgfplotsset{compat=1.18}
\definecolor{plotblue}{RGB}{58,111,185}
\definecolor{plotteal}{RGB}{0,128,128}
\definecolor{plotgreen}{RGB}{68,150,80}
\definecolor{plotorange}{RGB}{214,124,35}
\definecolor{plotred}{RGB}{190,64,64}

\newcolumntype{Y}{>{\raggedright\arraybackslash}X}
\newcommand{\method}{PRTE Agent\xspace}
\newcommand{\gym}{BioAgent Gym\xspace}
\newcommand{\bench}{BioWorkflow Bench\xspace}
\newcommand{\tactics}{BioWorkflow Tactics\xspace}

\title{Process-Reward Tactic Evolution for Long-Horizon Bioinformatics Workflows}

\author{
  Lingzhi Yang \\
  Stony Brook University
  \And
  Yubo Fan \\
  Vanderbilt University
  \And
  Song Wu \\
  Stony Brook University
  \And
  Gilchan Park \\
  Brookhaven National Lab
}

\begin{document}
\maketitle

\begin{abstract}
LLM agents can write code and call tools, but reliable bioinformatics work requires long-horizon interaction with workflow software, typed data objects, provenance, and biological checks. We study this setting through Galaxy workflow execution. The agent must explore task data, construct or adapt an executable workflow DAG, bind inputs and dataset collections, monitor execution, debug failures, and validate biological outputs. We propose \emph{Process-Reward Tactic Evolution}, a Galaxy-based training framework that turns verified workflow rollouts into reusable \tactics. During training, agents practice on curriculum-organized Galaxy tasks in \gym; process verifiers score workflow construction, software interaction, execution, and biological correctness; successful and failed traces are distilled into a tactic library. At inference, the trained executor, \method, uses this library to execute held-out \bench and BioAgent Bench tasks in isolated environments. The paper evaluates whether process-supervised tactic accumulation improves long-horizon bioinformatics workflow completion, biological correctness, and execution efficiency over no-memory and reflection-style baselines.

\end{abstract}

\section{Introduction}
\label{sec:introduction}

LLM agents are increasingly evaluated in long-horizon scientific settings where success depends on files, software state, tool calls, and execution traces rather than a single textual answer \citep{chen2025scienceagentbench,shen2026sciagentgym}. Bioinformatics workflow construction is a demanding instance: real analyses involve typed datasets, workflow graphs, provenance, biological assertions, and live execution on platforms such as Galaxy \citep{mitchener2025bixbench,fa2026bioagentbench}. Existing reflection, memory, skill, and prompt-evolution methods show that agents can improve by accumulating external process state \citep{shinn2023reflexion,zhao2024expel,wang2024awm,xu2025amem,agrawal2026gepa}, but they rarely train on realistic scientific workflow execution where intermediate software states are observable and consequential.

We ask how an agent can learn executable bioinformatics workflows from repeated, verifiable Galaxy interactions. Unlike biomedical QA, correctness is distributed across documentation lookup, input staging, datatype binding, workflow import, invocation, job completion, artifact collection, and biological validation. Useful learning must therefore be procedural: a later agent should recover tactics for schema inspection, collection wiring, graph repair, output accounting, and domain-specific workflow families, not only remember that a past task succeeded.

We propose Process-Reward Tactic Evolution, a Galaxy-based training framework for workflow agents. Agents practice on curriculum-organized tasks in \gym with public documentation, sandboxed notebooks, BioBlend, gxformat2, and usegalaxy execution backends \citep{iwc,bioblend,gxformat2}. Process verifiers score workflow construction, software interaction, execution, repair, leakage safety, and biological correctness \citep{sohn2026pra,lee2026metaharness,rosset2026verifiers}. A semantic-gradient updater converts successful and failed public trajectory segments into scoped tactics with applicability conditions, executable procedures, postconditions, failure signatures, and public evidence. At inference time, isolated workers retrieve the trained tactics and execute held-out workflow tasks.

We evaluate on \bench, a Galaxy benchmark derived from public workflow resources with 26 training tasks and 35 held-out test tasks \citep{iwc,alam2025prompttopipeline,cynthia2025galaxyretrieval}, and on BioAgent Bench as an external bioinformatics-agent benchmark \citep{fa2026bioagentbench}. Our contributions are to formulate Galaxy workflow construction as a long-horizon agent-training problem, introduce process-supervised tactic evolution with gated tactic-library growth, and show improved long/xlong BioWorkflow performance, biological correctness, and execution efficiency over no-memory and reflection/prompt-evolution baselines.

\section{Method}
\label{sec:method}

\subsection{Problem and Overview}

We study executable bioinformatics workflow construction as an agent-environment problem. Unlike short-answer biomedical QA, a Galaxy task is solved only when the agent builds, imports, runs, repairs, and validates a workflow through a real software backend. Formally, a task-output pair is
\begin{equation}
\begin{aligned}
  x&=(u,\mathcal I,D_{\mathrm{pub}},b),\\
  y&=(W,\hat{\mathcal D},A,\Pi,\tau,\mathbf r).
\end{aligned}
\end{equation}
Here \(u\) is the biological objective, \(\mathcal I\) is the input manifest, \(D_{\mathrm{pub}}\) is public documentation, and \(b\) contains backend constraints. The output includes an executable Galaxy workflow \(W\), produced datasets \(\hat{\mathcal D}\), biological assertions \(A\), provenance \(\Pi\), an execution trace \(\tau\), and process scores \(\mathbf r\). We use Galaxy because it offers maintained tools, importable workflow graphs, public IWC/GTN resources, and reproducible APIs \citep{iwc,bioblend,gxformat2}.

\begin{figure}[t]
\centering
\includegraphics[width=\columnwidth]{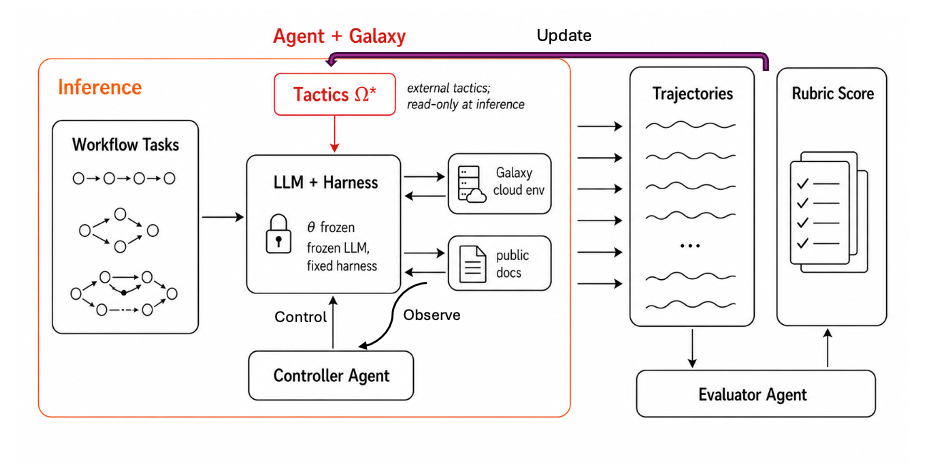}
\caption{Overview of Process-Reward Tactic Evolution. Training executes curriculum-organized Galaxy rollouts, scores them with process rewards, and updates an external tactic state. Inference retrieves the trained tactic library for isolated workers executing held-out Galaxy workflow tasks.}
\label{fig:method-overview}
\end{figure}

The workflow \(W=(V,E,\phi,\psi,\delta)\) is a typed DAG: nodes are tools, inputs, and outputs; edges carry datasets or collections; \(\phi\) gives tool identities; \(\psi\) stores parameters; and \(\delta\) stores datatype and collection constraints. Process-Reward Tactic Evolution (PRTE) keeps the execution environment stable across agents and improves an external tactic state \(\Omega\) from real Galaxy rollouts. Figure~\ref{fig:method-overview} shows the training and inference split.

\subsection{Tactic State}

PRTE represents learned experience as a tactic library rather than a single reflection string. This follows memory and skill-library agents, but specializes the stored units to verifier-backed scientific workflow procedures \citep{wang2024awm,zhou2025memento,liu2026gos,liang2026skillnet}. We write
\begin{equation}
  \Omega=(\mathcal L,G^\Omega,\rho),
\end{equation}
where \(\mathcal L\) is a set of tactic files, \(G^\Omega\) is a dependency graph, and \(\rho\) is a retrieval function. A tactic states when it applies, what operation to perform, what postcondition should hold, which failure signature it repairs, and what public evidence supports it. Typical tactics cover schema inspection, paired-collection wiring, gxformat2 checking, Galaxy invocation repair, artifact accounting, and task-family biological checks.

\subsection{Process-Supervised Updates}

Training proceeds over curriculum batches \(S_k\). Each batch uses one tactic snapshot, runs isolated Galaxy rollouts, and then updates the tactic state:
\begin{equation}
\begin{aligned}
  \mathcal B_k&=\{\tau_{x,j}:x\in S_k,\;j=1,\ldots,g_k(x)\},\\
  \Omega_{k+1}&=U(\Omega_k;\{(\tau_i,\mathbf r_i)\}_{i\in\mathcal B_k}).
\end{aligned}
\end{equation}
The curriculum starts with small workflows, documentation lookup, local checks, and importable graphs, then adds longer dependency chains, live invocation, queue behavior, artifact recovery, and biological validation.

Following process-reward and verifier-harness work, the updater receives intermediate evidence rather than only final success \citep{sohn2026pra,lee2026metaharness,rosset2026verifiers}. Each trajectory is scored as
\begin{equation}
\begin{aligned}
\mathbf r(\tau)=(&r_{\mathrm{doc}},r_{\mathrm{chk}},r_{\mathrm{imp}},
r_{\mathrm{inv}},r_{\mathrm{job}},\\
&r_{\mathrm{art}},r_{\mathrm{bio}},r_{\mathrm{rep}},
r_{\mathrm{life}},r_{\mathrm{safe}}).
\end{aligned}
\end{equation}
These channels measure documentation use, local validation, import, invocation, job completion, artifact collection, biological correctness, repair progress, lifecycle discipline, and leakage safety. The scalar training signal and group-relative credit are
\begin{equation}
  R_i=\frac{\sum_j w_j\,\mathrm{clip}_{[0,1]}(r_{ij})}{\sum_jw_j},
  \qquad
  A_i=\frac{R_i-\mu_{\mathcal B}}{\sigma_{\mathcal B}+\epsilon}.
\end{equation}

The update \(U\) has three stages. A reviewer extracts public positive and negative evidence from each trajectory. A semantic-gradient updater compares successful and failed excerpts to propose tactic patches. A gate accepts a patch only if it has applicability, executable procedure, verification check, failure signature, and public evidence. Hidden gold workflows and private assertions are never written into \(\Omega\).

\subsection{Tactic-Aware Inference}

Held-out inference freezes the trained tactic state. A worker retrieves a budgeted tactic bundle
\begin{equation}
  B_x=\mathrm{Hydrate}\!\left(\mathrm{Closure}(\rho(x,\mathcal L),G^\Omega),B_{\max}\right),
\end{equation}
then executes the task in an isolated notebook and Galaxy account. It stages data, writes or adapts workflow artifacts, runs local checks, imports and invokes the graph, polls jobs, repairs allowed failures, collects outputs, and submits evidence. A controller monitors usage windows, authentication, backend instability, resumes, timeouts, and cleanup, but it does not choose tools, edit graphs, or provide biological answers.

\section{Experiments}
\label{sec:experiments}

\subsection{Tasks and Environment}

We evaluate in a Galaxy-centered bioinformatics environment. Each run uses a local sandbox with Python notebooks, BioBlend, gxformat2, task manifests, public documentation, and a usegalaxy execution backend. The agent must stage inputs, construct or adapt a workflow, import and invoke it, monitor execution, collect artifacts, and write evaluator-readable evidence. The training split contains 26 workflow tasks and the held-out split contains 35 tasks, which we call \texttt{test35}. The held-out set is balanced across small, medium, long, and xlong workflow sizes. Appendix~\ref{app:bench-preparation} lists the training and test tasks and the public Galaxy documentation used by the agents.

\subsection{Benchmarks}

We use two complementary benchmarks. BioWorkflow Bench is our main benchmark for long-horizon Galaxy workflow construction. It contains executable workflow tasks derived from public workflow resources and records graph statistics such as tool count and workflow depth. BioAgent Bench is an external 10-task bioinformatics-agent benchmark \citep{fa2026bioagentbench}; it is shorter, but it provides an outside point of comparison for result matching, final-result completion, and step-level completion. The benchmark statistics in Table~\ref{tab:two-benchmark-statistics} make this contrast explicit: our held-out split contains deeper and more structurally varied workflow graphs, while BioAgent Bench anchors the comparison to an existing external suite.

\begin{table}[t]
\centering
\scriptsize
\resizebox{\columnwidth}{!}{%
\begin{tabular}{@{}llrrrr@{}}
\toprule
Benchmark & Split & \multicolumn{1}{r}{Tasks} & \multicolumn{1}{r}{Tools} & \multicolumn{1}{r}{Avg. tools} & \multicolumn{1}{r}{Avg. depth} \\
\midrule
\multirow{3}{*}{BioWorkflow Bench}
  & train26 & 26 & 322 & 12.38 & 10.62 \\
  & test35  & 35 & 455 & 13.00 & 11.34 \\
  & total61 & 61 & 777 & 12.74 & 11.03 \\
BioAgent Bench & full & 10 & 62 & 6.20 & 5.80 \\
\bottomrule
\end{tabular}
}

\caption{Benchmark statistics. BioWorkflow Bench \texttt{test35} refers to the final 35-task held-out complexity-balanced test split; BioAgent Bench \texttt{full} refers to the adopted 10-task external benchmark set. Metric definitions and task-level details are provided in Appendix~\ref{app:bench-preparation}.}
\label{tab:two-benchmark-statistics}
\end{table}

\subsection{Evaluated Agents}

We evaluate eight agents on BioWorkflow Bench \texttt{test35}. The no-training group includes ReAct, MultiAgent, and Agent Harness/Biomni. The training group includes Reflexion, Agent Workflow Memory, Agentic Memory, GEPA-style prompt evolution, and \method. All agents use the same model family, sandbox interface, public task bundles, and evaluator. For BioAgent Bench, we evaluate the four agents for which comparable task-level evidence is available: ReAct, MultiAgent, Agent Harness/Biomni, and \method. For converted BioWorkflow rows, we use the same quick10-to-test35 conversion protocol documented with the result tables; actual test35 rows are explicitly marked in the captions.

\subsection{Metrics}

BioWorkflow Bench reports workflow completion, biological correctness, and a weighted final score:
\begin{equation}
  \mathrm{Final}=0.6\cdot\mathrm{Workflow}+0.4\cdot\mathrm{Biology}.
\end{equation}
Workflow completion measures whether the agent constructs and executes the requested Galaxy workflow process. Biology measures whether the outputs and assertions match the intended biological analysis. We also report token consumption, API rounds, execution time, and milestone progress for an xlong Hi-C case study.

\section{Main Results}
\label{sec:results}

\subsection{Overall BioWorkflow Performance}

\begin{figure}[t]
\centering
\definecolor{notrainA}{RGB}{230,146,42}
\definecolor{notrainB}{RGB}{242,177,73}
\definecolor{notrainC}{RGB}{247,204,111}
\definecolor{trainA}{RGB}{99,171,104}
\definecolor{trainB}{RGB}{61,145,86}
\definecolor{trainC}{RGB}{136,192,116}
\definecolor{gepaBlue}{RGB}{70,124,191}
\definecolor{prtePurple}{RGB}{137,89,190}
\begin{tikzpicture}
\begin{axis}[
    width=\columnwidth,
    height=0.68\columnwidth,
    ybar,
    bar width=1.75pt,
    ymin=0,
    ymax=0.90,
    symbolic x coords={Small,Medium,Long,XLong,Overall},
    xtick=data,
    enlarge x limits=0.18,
    ylabel={BioWorkflow Bench final score},
    ylabel style={font=\scriptsize},
    xticklabel style={font=\scriptsize},
    yticklabel style={font=\scriptsize},
    ymajorgrids,
    grid style={draw=gray!18},
    axis line style={draw=gray!60},
    tick style={draw=none},
    major tick length=0pt,
    legend columns=2,
    legend style={
      at={(0.985,0.985)},
      anchor=north east,
      draw=gray!25,
      fill=white,
      fill opacity=0.92,
      text opacity=1,
      font=\fontsize{4.8}{5.5}\selectfont,
      inner sep=0.65pt,
      row sep=-0.25pt,
      column sep=0.55pt,
      nodes={inner sep=0.25pt,text depth=0.25ex},
      /tikz/every even column/.append style={column sep=0.75pt}
    },
    legend image code/.code={
      \draw[#1,draw=none] (0cm,-0.03cm) rectangle (0.09cm,0.03cm);
    },
]
\addplot+[draw=none, fill=notrainA] coordinates {
  (Small,0.5867) (Medium,0.5362) (Long,0.2592) (XLong,0.1934) (Overall,0.3663)
};
\addplot+[draw=none, fill=notrainB] coordinates {
  (Small,0.5635) (Medium,0.5164) (Long,0.2177) (XLong,0.2117) (Overall,0.3507)
};
\addplot+[draw=none, fill=notrainC] coordinates {
  (Small,0.5723) (Medium,0.5716) (Long,0.2386) (XLong,0.2779) (Overall,0.3926)
};
\addplot+[draw=none, fill=trainA] coordinates {
  (Small,0.7071) (Medium,0.5522) (Long,0.2477) (XLong,0.2841) (Overall,0.4107)
};
\addplot+[draw=none, fill=trainB] coordinates {
  (Small,0.7650) (Medium,0.6357) (Long,0.2799) (XLong,0.3071) (Overall,0.4586)
};
\addplot+[draw=none, fill=trainC] coordinates {
  (Small,0.7168) (Medium,0.5961) (Long,0.3096) (XLong,0.2858) (Overall,0.4428)
};
\addplot+[draw=none, fill=gepaBlue] coordinates {
  (Small,0.8420) (Medium,0.6977) (Long,0.3059) (XLong,0.3090) (Overall,0.4953)
};
\addplot+[draw=none, fill=prtePurple] coordinates {
  (Small,0.7753) (Medium,0.6565) (Long,0.3182) (XLong,0.3922) (Overall,0.5013)
};
\legend{ReAct, MultiAgent, Agent Harness, Reflexion, AWM, Agentic-Memory, GEPA, PRTE Agent}
\end{axis}
\end{tikzpicture}
\caption{BioWorkflow Bench test35 final weighted scores by complexity group and overall. The three no-training agents use orange/yellow shades; Reflexion, AWM, and Agentic-Memory use green shades; GEPA uses blue; and the Process-Reward Tactic Evolution agent is denoted as PRTE Agent in purple.}
\label{fig:bioworkflow-test35-by-group}
\end{figure}
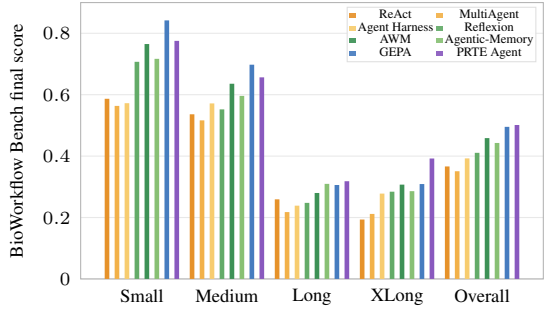

Memory and training help on \bench, but their value depends on whether the learned state can support long workflow execution. GEPA obtains the highest small and medium scores, where a strong global instruction can cover many tasks. The pattern reverses as workflow horizon grows: \method obtains the best long, xlong, and overall final scores, as shown in Figure~\ref{fig:bioworkflow-test35-by-group}. Its overall score is 0.5013, compared with 0.4953 for GEPA, 0.4586 for Agent Workflow Memory, and 0.3926 for Agent Harness/Biomni.

\begin{table*}[t!]
\centering
\scriptsize
\setlength{\tabcolsep}{2.2pt}
\renewcommand{\arraystretch}{1.08}
\resizebox{\textwidth}{!}{%
\begin{tabular}{@{}llrrrrrrrrrrrrrrr@{}}
\toprule
\multirow{2}{*}{Group} & \multirow{2}{*}{Agent}
& \multicolumn{3}{c}{Small ($n=5$)}
& \multicolumn{3}{c}{Medium ($n=10$)}
& \multicolumn{3}{c}{Long ($n=10$)}
& \multicolumn{3}{c}{XLong ($n=10$)}
& \multicolumn{3}{c}{Overall ($n=35$)} \\
\cmidrule(lr){3-5}\cmidrule(lr){6-8}\cmidrule(lr){9-11}\cmidrule(lr){12-14}\cmidrule(l){15-17}
& & Workflow & Biology & Final
& Workflow & Biology & Final
& Workflow & Biology & Final
& Workflow & Biology & Final
& Workflow & Biology & Final \\
\midrule
No training & ReAct
& 0.7691 & 0.3132 & 0.5867
& 0.6984 & 0.2928 & 0.5362
& 0.2794 & 0.2287 & 0.2592
& 0.2233 & 0.1485 & 0.1934
& 0.4531 & 0.2362 & 0.3663 \\
No training & MultiAgent
& 0.7530 & 0.2793 & 0.5635
& 0.7210 & 0.2096 & 0.5164
& 0.2958 & 0.1006 & 0.2177
& 0.2650 & 0.1317 & 0.2117
& 0.4738 & 0.1662 & 0.3507 \\
No training & Agent Harness
& 0.7192 & 0.3519 & 0.5723
& 0.7122 & 0.3606 & 0.5716
& 0.2998 & 0.1468 & 0.2386
& 0.2270 & 0.3543 & 0.2779
& 0.4567 & 0.2965 & 0.3926 \\
\midrule
Training & Reflexion
& 0.8595 & 0.4785 & 0.7071
& 0.7565 & 0.2458 & 0.5522
& 0.3564 & 0.0846 & 0.2477
& \textbf{0.3658} & 0.1614 & 0.2841
& 0.5453 & 0.2089 & 0.4107 \\
Training & Agent Workflow Memory
& 0.8489 & 0.6391 & 0.7650
& 0.7221 & 0.5062 & 0.6357
& 0.2957 & \textbf{0.2561} & 0.2799
& 0.3039 & 0.3119 & 0.3071
& 0.4989 & 0.3982 & 0.4586 \\
Training & Agentic Memory
& 0.8590 & 0.5034 & 0.7168
& 0.7465 & 0.3705 & 0.5961
& \textbf{0.3810} & 0.2026 & 0.3096
& 0.3539 & 0.1836 & 0.2858
& \textbf{0.5460} & 0.2881 & 0.4428 \\
Training & GEPA
& \textbf{0.9383} & \textbf{0.6976} & \textbf{0.8420}
& \textbf{0.7648} & \textbf{0.5970} & \textbf{0.6977}
& 0.3620 & 0.2217 & 0.3059
& 0.2610 & 0.3808 & 0.3090
& 0.5305 & 0.4424 & 0.4953 \\
Training & PRTE Agent
& 0.8271 & \textbf{0.6976} & 0.7753
& 0.7206 & 0.5603 & 0.6565
& 0.3764 & 0.2309 & \textbf{0.3182}
& 0.3535 & \textbf{0.4502} & \textbf{0.3922}
& 0.5326 & \textbf{0.4544} & \textbf{0.5013} \\
\bottomrule
\end{tabular}
}
\caption{Detailed BioWorkflow Bench test35 results for all evaluated agents. Each complexity group reports workflow completion, biological correctness, and the weighted final score. Bold marks the best value in each metric column, with ties bolded together. PRTE Agent denotes the Process-Reward Tactic Evolution agent. GEPA and PRTE Agent are actual test35 runs; the other BioWorkflow agents are converted test35 estimates.}
\label{tab:bioworkflow-test35-detailed}
\end{table*}

The overall gain is not uniform; it is concentrated in the hardest workflow regimes. The detailed decomposition in Table~\ref{tab:bioworkflow-test35-detailed} shows that \method improves over GEPA on long workflows (0.3182 vs. 0.3059 final) and xlong workflows (0.3922 vs. 0.3090 final). The xlong gain comes mainly from biological correctness, where \method reaches 0.4502 while GEPA reaches 0.3808. This supports the intended role of tactics: they preserve platform procedures while leaving more model context for biological reasoning and output validation.

\subsection{External BioAgent Bench Comparison}

\begin{table}[t!]
\centering
\small
\setlength{\tabcolsep}{3.2pt}
\renewcommand{\arraystretch}{1.05}
\begin{tabularx}{\linewidth}{@{}lrrrr@{}}
\toprule
Agent & Match & Final & Step & Combined \\
\midrule
ReAct & 2/10 & 9/10 & 0.8600 & 0.6533 \\
MultiAgent & 5/10 & 9/10 & 0.9260 & 0.7753 \\
Agent Harness & \textbf{8/10} & \textbf{10/10} & 0.9500 & 0.9167 \\
PRTE Agent & \textbf{8/10} & \textbf{10/10} & \textbf{0.9667} & \textbf{0.9222} \\
\bottomrule
\end{tabularx}
\caption{BioAgent Bench results on the external 10-task benchmark. Match is the result-match rate, Final is the final-result-reached rate, Step is the average step-level completion score, and Combined is their mean. Agent Harness corresponds to Biomni; PRTE Agent denotes the Process-Reward Tactic Evolution agent.}
\label{tab:bioagent-bench-results}
\end{table}

The learned tactic state also transfers to a shorter external benchmark. On BioAgent Bench, \method reaches 10/10 final-result completion, 8/10 result match, and the highest average step score (Table~\ref{tab:bioagent-bench-results}). Agent Harness/Biomni also reaches 10/10 final completion and 8/10 result match, while \method has a slightly higher step-level score and combined mean. This confirms that the gains are not only an artifact of our own benchmark construction; the same process memory remains competitive on an external bioinformatics-agent evaluation.

\subsection{Ablation Study}

\begin{figure}[t!]
\centering
\definecolor{gepaBlue}{RGB}{70,124,191}
\definecolor{noProcessGreen}{RGB}{99,171,104}
\definecolor{prtePurple}{RGB}{137,89,190}
\begin{tikzpicture}
\begin{axis}[
    width=\columnwidth,
    height=0.70\columnwidth,
    ymin=0.38,
    ymax=0.52,
    symbolic x coords={empty,batch1,batch2,batch3,batch4},
    xtick=data,
    xlabel={Checkpoint},
    ylabel={Overall final score},
    xlabel style={font=\scriptsize},
    ylabel style={font=\scriptsize},
    xticklabel style={font=\fontsize{7.2}{8.0}\selectfont},
    yticklabel style={font=\scriptsize},
    ymajorgrids,
    grid style={draw=gray!18},
    axis line style={draw=gray!60},
    tick style={draw=gray!50},
    line width=0.9pt,
    mark size=1.8pt,
    legend columns=3,
    legend style={
      at={(0.50,1.04)},
      anchor=south,
      draw=gray!25,
      fill=white,
      fill opacity=0.92,
      text opacity=1,
      font=\scriptsize,
      inner sep=1.3pt,
      row sep=-0.5pt
    },
]
\addplot+[color=gepaBlue, mark=square*, mark options={fill=gepaBlue}] coordinates {
  (empty,0.3982) (batch1,0.4508) (batch2,0.4821) (batch3,0.4237) (batch4,0.4237)
};
\addlegendentry{GEPA}
\addplot+[color=noProcessGreen, mark=triangle*, mark options={fill=noProcessGreen}] coordinates {
  (empty,0.4483) (batch1,0.4842) (batch2,0.4686) (batch3,0.4569) (batch4,0.4770)
};
\addlegendentry{No Process}
\addplot+[color=prtePurple, mark=*, mark options={fill=prtePurple}] coordinates {
  (empty,0.4483) (batch1,0.4740) (batch2,0.4889) (batch3,0.4942) (batch4,0.5013)
};
\addlegendentry{Process Reward}
\end{axis}
\end{tikzpicture}
\caption{Ablation training curve across checkpoints. The plotted value is the overall BioWorkflow Bench final score for each configuration.}
\label{fig:ablation-training-curve}
\end{figure}
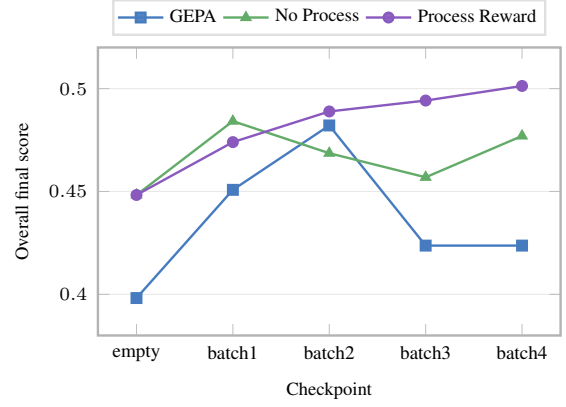
\begin{table}[t!]
\centering
\scriptsize
\setlength{\tabcolsep}{2.8pt}
\renewcommand{\arraystretch}{1.08}
\resizebox{\linewidth}{!}{%
\begin{tabular}{@{}llll@{}}
\toprule
Checkpoint & GEPA & No Process & Process Reward \\
\midrule
empty & 0.4631 / 0.3024 / 0.3982 & 0.4799 / 0.4008 / 0.4483 & 0.4799 / 0.4008 / 0.4483 \\
batch1 & 0.5571 / 0.2912 / 0.4508 & 0.5238 / 0.4248 / 0.4842 & 0.5305 / 0.3893 / 0.4740 \\
batch2 & 0.5504 / 0.3796 / 0.4821 & 0.5167 / 0.3964 / 0.4686 & 0.5546 / 0.3904 / 0.4889 \\
batch3 & 0.5481 / 0.2372 / 0.4237 & 0.5258 / 0.3535 / 0.4569 & 0.5619 / 0.3925 / 0.4942 \\
batch4 & 0.5481 / 0.2372 / 0.4237 & 0.5008 / 0.4415 / 0.4770 & 0.5326 / 0.4544 / 0.5013 \\
\bottomrule
\end{tabular}
}
\caption{Training-curve ablation results across checkpoints. Each cell reports Workflow / Biology / Final. Process Reward is the PRTE training configuration with process-reward-guided tactic evolution.}
\label{tab:ablation-training-curve}
\end{table}

Process reward is the part of training that turns repeated rollouts into usable procedural memory. Across checkpoints, outcome-only training improves after the first batch but does not keep improving; its final score ends at 0.4770. GEPA remains strong but flat after its selected seed instruction. The full process-reward agent improves more steadily and finishes at 0.5013. Figure~\ref{fig:ablation-training-curve} and Table~\ref{tab:ablation-training-curve} show this trajectory from the same checkpoint scores. The ablation result is therefore structural: repeated outcome labels alone are insufficient for long workflow learning, while process signals provide the updater with enough evidence to write reusable tactics.

\begin{table}[t!]
\centering
\small
\setlength{\tabcolsep}{4.0pt}
\renewcommand{\arraystretch}{1.08}
\begin{tabular}{@{}lrrrr@{}}
\toprule
\multirow{2}{*}{Agent} & \multicolumn{2}{c}{Init/Empty} & \multicolumn{2}{c}{After Batch4} \\
\cmidrule(lr){2-3}\cmidrule(l){4-5}
& Files & Tokens & Files & Tokens \\
\midrule
GEPA & 1 & 223 & 1 & 852 \\
No Process & 6 & 6,669 & 8 & 11,039 \\
Process Reward & 6 & 6,669 & 26 & 33,700 \\
\bottomrule
\end{tabular}
\caption{Tactics-library size before training and after batch4. GEPA mutates a single instruction file, while No Process and Process Reward maintain tactic Markdown files. Token counts use the same approximate convention as the training analysis.}
\label{tab:tactics-statistics}
\end{table}

The checkpoint curves diverge because the training variants write different kinds of state. GEPA evolves a single instruction. The no-process library grows modestly from 6 to 8 tactic files. The process-reward library grows to 26 tactic files and 33.7K approximate tokens (Table~\ref{tab:tactics-statistics}). The qualitative logs in Appendix~\ref{app:experimental-comparisons} show that these additional files are not merely longer reflections: they include repair procedures and task-family contracts for recurring workflow families such as RNA-seq, Hi-C, pseudobulk analysis, MAG generation, HyPhy comparison, assembly cleanup, and chromatin peak QC. Thus process reward changes the kind of memory written, not only its size.

\subsection{Resource Efficiency}

\begin{table}[t!]
\centering
\scriptsize
\setlength{\tabcolsep}{3.0pt}
\renewcommand{\arraystretch}{1.05}
\resizebox{\linewidth}{!}{%
\begin{tabular}{@{}lccc@{}}
\toprule
Agent & Token & API Rounds & Execution Time \\
\midrule
ReAct & 2662.939M / 76.084M & 1,430.0 / 40.9 & 84.48h / 2.41h \\
MultiAgent & 8514.015M / 243.258M & 4,973.3 / 142.1 & 153.84h / 4.40h \\
Agent Harness / Biomni & 2780.308M / 79.437M & 1,426.7 / 40.8 & 63.17h / 1.80h \\
Reflexion & 689.846M / 19.710M & 6,156.7 / 175.9 & 23.16h / 0.66h \\
Agent Workflow Memory & 1474.397M / 42.126M & 8,220.0 / 234.9 & 65.55h / 1.87h \\
Agentic Memory & 739.059M / 21.116M & 3,798.3 / 108.5 & 26.81h / 0.77h \\
GEPA & 765.597M / 21.874M & 5,551.0 / 158.6 & 25.82h / 0.74h \\
PRTE Agent & 963.454M / 27.527M & 7,746.0 / 221.3 & 33.37h / 0.95h \\
\bottomrule
\end{tabular}
}
\caption{Test35 resource consumption. Each cell reports total consumption followed by the average per task. Token counts are in millions of tokens and execution time is in hours.}
\label{tab:test35-token-consumption}
\end{table}

Tactics also change the cost profile of held-out execution. No-training agents spend very large token budgets because they rediscover platform procedures during every task. MultiAgent is the most expensive, with 8.514B total tokens and 4,973 API rounds. \method uses 963.454M total tokens on the held-out tasks, averaging 27.527M per task (Table~\ref{tab:test35-token-consumption}). Training agents can have substantial API-round counts because they perform detailed tool and workflow interactions, but the important comparison is that tactic-guided execution avoids the very large token growth seen in no-training multi-agent runs.

\begin{figure}[t!]
\centering
\definecolor{notrainA}{RGB}{230,146,42}
\definecolor{notrainB}{RGB}{242,177,73}
\definecolor{notrainC}{RGB}{247,204,111}
\definecolor{trainA}{RGB}{99,171,104}
\definecolor{trainB}{RGB}{61,145,86}
\definecolor{trainC}{RGB}{136,192,116}
\definecolor{gepaBlue}{RGB}{70,124,191}
\definecolor{prtePurple}{RGB}{137,89,190}
\begin{tikzpicture}
\begin{axis}[
    width=\columnwidth,
    height=0.72\columnwidth,
    xmin=0,
    xmax=35,
    ymin=0,
    ymax=3000,
    xlabel={Evaluated Tasks},
    ylabel={Cum. Tokens (M)},
    xlabel style={font=\scriptsize},
    ylabel style={font=\scriptsize},
    xtick={0,5,10,15,20,25,30,35},
    ytick={0,500,1000,1500,2000,2500,3000},
    xticklabel style={font=\scriptsize},
    yticklabel style={font=\scriptsize},
    ymajorgrids,
    xmajorgrids,
    grid style={draw=gray!16},
    axis line style={draw=gray!60},
    tick style={draw=gray!50},
    line width=0.95pt,
    mark size=0.65pt,
    legend columns=2,
    legend style={
      at={(0.98,0.04)},
      anchor=south east,
      draw=gray!25,
      fill=white,
      fill opacity=0.93,
      text opacity=1,
      font=\fontsize{5.2}{5.9}\selectfont,
      inner sep=1.0pt,
      row sep=-0.9pt,
      column sep=1.2pt
    },
]
\addplot+[color=notrainA, mark=none] coordinates {
  (0,0.000) (1,76.084) (2,152.168) (3,228.252) (4,304.336) (5,380.420)
  (6,456.504) (7,532.588) (8,608.672) (9,684.756) (10,760.840)
  (11,836.924) (12,913.008) (13,989.092) (14,1065.176) (15,1141.260)
  (16,1217.344) (17,1293.428) (18,1369.511) (19,1445.595) (20,1521.679)
  (21,1597.763) (22,1673.847) (23,1749.931) (24,1826.015) (25,1902.099)
  (26,1978.183) (27,2054.267) (28,2130.351) (29,2206.435) (30,2282.519)
  (31,2358.603) (32,2434.687) (33,2510.771) (34,2586.855) (35,2662.939)
};
\addlegendentry{ReAct}
\addplot+[color=notrainB, mark=none] coordinates {
  (0,0.000) (1,145.374) (2,710.648) (3,1000.360) (4,1035.107)
  (5,1094.371) (6,1343.596) (7,1613.148) (8,1708.668) (9,1948.053)
  (10,2347.477) (11,2595.942) (12,2844.407)
};
\addlegendentry{MultiAgent}
\addplot+[color=notrainC, mark=none] coordinates {
  (0,0.000) (1,51.387) (2,119.765) (3,229.840) (4,266.592) (5,298.655)
  (6,525.380) (7,533.683) (8,572.826) (9,615.110) (10,692.202)
  (11,744.374) (12,796.547) (13,906.621) (14,1016.696) (15,1126.771)
  (16,1236.846) (17,1346.920) (18,1456.995) (19,1567.070) (20,1677.144)
  (21,1787.219) (22,1876.249) (23,1965.280) (24,2054.310) (25,2143.341)
  (26,2232.371) (27,2321.401) (28,2410.432) (29,2463.271) (30,2516.111)
  (31,2568.950) (32,2621.790) (33,2674.629) (34,2727.469) (35,2780.308)
};
\addlegendentry{Harness}
\addplot+[color=trainA, mark=none] coordinates {
  (0,2869.512) (1,2881.274) (2,2896.228) (3,2924.066) (4,2935.289)
  (5,2951.907) (6,2968.525) (7,2985.143)
};
\addlegendentry{Reflexion}
\addplot+[color=trainB, mark=none] coordinates {
  (0,626.202) (1,668.328) (2,710.453) (3,752.579) (4,794.705)
  (5,836.830) (6,878.956) (7,921.081) (8,963.207) (9,1005.333)
  (10,1047.458) (11,1089.584) (12,1131.710) (13,1173.835) (14,1215.961)
  (15,1258.086) (16,1300.212) (17,1342.338) (18,1384.463) (19,1426.589)
  (20,1468.715) (21,1510.840) (22,1552.966) (23,1595.091) (24,1637.217)
  (25,1679.343) (26,1721.468) (27,1763.594) (28,1805.720) (29,1847.845)
  (30,1889.971) (31,1932.096) (32,1974.222) (33,2016.348) (34,2058.473)
  (35,2100.599)
};
\addlegendentry{AWM}
\addplot+[color=trainC, mark=none] coordinates {
  (0,824.537) (1,845.653) (2,866.769) (3,887.885) (4,909.001)
  (5,930.117) (6,951.233) (7,972.349) (8,993.465) (9,1014.581)
  (10,1035.697) (11,1056.813) (12,1077.929) (13,1099.045) (14,1120.161)
  (15,1141.277) (16,1162.393) (17,1183.509) (18,1204.624) (19,1225.740)
  (20,1246.856) (21,1267.972) (22,1289.088) (23,1310.204) (24,1331.320)
  (25,1352.436) (26,1373.552) (27,1394.668) (28,1415.784) (29,1436.900)
  (30,1458.016) (31,1479.132) (32,1500.248) (33,1521.364) (34,1542.480)
  (35,1563.596)
};
\addlegendentry{A-Memory}
\addplot+[color=gepaBlue, mark=none] coordinates {
  (0,1956.328) (1,1964.813) (2,1973.833) (3,1994.385) (4,2002.539)
  (5,2025.616) (6,2038.460) (7,2060.703) (8,2090.715) (9,2100.716)
  (10,2151.562) (11,2163.444) (12,2193.573) (13,2203.068) (14,2217.295)
  (15,2230.278) (16,2247.674) (17,2252.464) (18,2282.942) (19,2302.660)
  (20,2328.403) (21,2359.791) (22,2370.523) (23,2407.420) (24,2434.375)
  (25,2464.825) (26,2490.006) (27,2515.427) (28,2546.490) (29,2559.409)
  (30,2573.699) (31,2582.101) (32,2628.770) (33,2654.296) (34,2694.830)
  (35,2721.925)
};
\addlegendentry{GEPA}
\addplot+[color=prtePurple, mark=none] coordinates {
  (0,880.086) (1,889.638) (2,905.114) (3,946.011) (4,956.593)
  (5,977.413) (6,998.587) (7,1026.430) (8,1051.195) (9,1081.084)
  (10,1113.102) (11,1151.372) (12,1163.282) (13,1186.896) (14,1198.336)
  (15,1216.346) (16,1247.539) (17,1260.052) (18,1300.426) (19,1315.792)
  (20,1342.791) (21,1374.527) (22,1400.951) (23,1451.795) (24,1481.960)
  (25,1523.056) (26,1563.965) (27,1601.259) (28,1625.627) (29,1663.373)
  (30,1684.746) (31,1703.221) (32,1724.727) (33,1752.884) (34,1797.996)
  (35,1843.540)
};
\addlegendentry{PRTE}
\end{axis}
\end{tikzpicture}
\caption{Cumulative token consumption over the ordered test35 tasks. Training agents include their fixed training cost at task count 0; curves stop after crossing the 3B-token plotting range.}
\label{fig:token-efficiency-curve}
\end{figure}
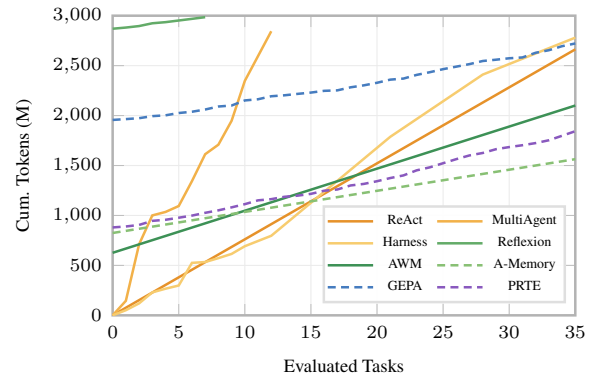

The amortization point matters because training is a fixed upfront cost. For a single short task, training may be unnecessary; for repeated long-horizon workflows in the same software environment, reusable tactics amortize their cost. Figure~\ref{fig:token-efficiency-curve} plots this cumulative token trajectory over ordered held-out tasks. This is the practical regime for bioinformatics platforms, where scientists repeatedly use the same infrastructure, datatypes, wrappers, and debugging patterns across different analyses.

\subsection{XLong Case Study}

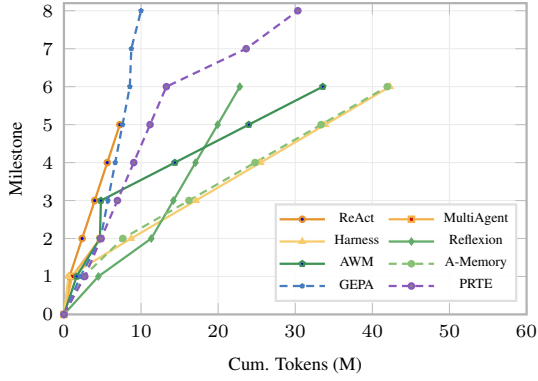
\begin{figure}[t!]
\centering
\definecolor{notrainA}{RGB}{230,146,42}
\definecolor{notrainB}{RGB}{242,177,73}
\definecolor{notrainC}{RGB}{247,204,111}
\definecolor{trainA}{RGB}{99,171,104}
\definecolor{trainB}{RGB}{61,145,86}
\definecolor{trainC}{RGB}{136,192,116}
\definecolor{gepaBlue}{RGB}{70,124,191}
\definecolor{prtePurple}{RGB}{137,89,190}
\begin{tikzpicture}
\begin{axis}[
    width=\columnwidth,
    height=0.74\columnwidth,
    xmin=0,
    xmax=60,
    ymin=0,
    ymax=8.2,
    xlabel={Cum. Tokens (M)},
    ylabel={Milestone},
    xlabel style={font=\scriptsize},
    ylabel style={font=\scriptsize},
    xtick={0,10,20,30,40,50,60},
    ytick={0,1,2,3,4,5,6,7,8},
    xticklabel style={font=\scriptsize},
    yticklabel style={font=\scriptsize},
    ymajorgrids,
    xmajorgrids,
    grid style={draw=gray!16},
    axis line style={draw=gray!60},
    tick style={draw=gray!50},
    line width=0.85pt,
    mark size=1.0pt,
    legend columns=2,
    legend style={
      at={(0.98,0.04)},
      anchor=south east,
      draw=gray!25,
      fill=white,
      fill opacity=0.93,
      text opacity=1,
      font=\fontsize{5.2}{5.9}\selectfont,
      inner sep=1.0pt,
      row sep=-0.8pt,
      column sep=1.2pt
    },
]
\addplot+[color=notrainA, mark=*] coordinates {(0.000,0) (0.732,1) (2.362,2) (3.992,3) (5.621,4) (7.251,5)};
\addlegendentry{ReAct}
\addplot+[color=notrainB, mark=square*] coordinates {(0.000,0) (1.047,1) (4.727,2)};
\addlegendentry{MultiAgent}
\addplot+[color=notrainC, mark=triangle*] coordinates {(0.000,0) (0.597,1) (8.713,2) (17.106,3) (25.499,4) (33.891,5) (42.284,6)};
\addlegendentry{Harness}
\addplot+[color=trainA, mark=diamond*] coordinates {(0.000,0) (4.466,1) (11.325,2) (14.196,3) (17.068,4) (19.939,5) (22.810,6)};
\addlegendentry{Reflexion}
\addplot+[color=trainB, mark=pentagon*] coordinates {(0.000,0) (1.641,1) (4.669,2) (4.796,3) (14.387,4) (23.979,5) (33.570,6)};
\addlegendentry{AWM}
\addplot+[color=trainC, mark=otimes*] coordinates {(0.000,0) (2.545,1) (7.636,2) (16.210,3) (24.784,4) (33.359,5) (41.933,6)};
\addlegendentry{A-Memory}
\addplot+[color=gepaBlue, mark=star] coordinates {(0.000,0) (2.136,1) (4.758,2) (5.704,3) (6.651,4) (7.598,5) (8.544,6) (8.739,7) (10.001,8)};
\addlegendentry{GEPA}
\addplot+[color=prtePurple, mark=*] coordinates {(0.000,0) (2.704,1) (4.821,2) (6.937,3) (9.054,4) (11.170,5) (13.287,6) (23.634,7) (30.336,8)};
\addlegendentry{PRTE}
\end{axis}
\end{tikzpicture}
\caption{Milestone progress on the representative xlong Hi-C task. Each curve shows the cumulative task-execution tokens needed to reach each milestone; training cost is excluded. Curves are truncated at the 60M-token plotting boundary.}
\label{fig:xlong-milestone-efficiency}
\end{figure}

The representative xlong Hi-C case shows why aggregate token use alone is not a sufficient efficiency metric. Several agents stop before graph success even after large token use, so long-horizon progress is not proportional to total reasoning effort. Figure~\ref{fig:xlong-milestone-efficiency} makes this visible by plotting milestone progress, from initialization through data preparation, notebook creation, Galaxy response, tool-format discovery, graph upload, successful graph execution, and final submission. Training cost is excluded so the figure measures only task execution. GEPA reaches all milestones with very low task-token cost on this case, while \method also reaches final submission and maintains a more complete execution trajectory than incomplete baselines. The appendix gives the tactics-to-trace comparison and the final Galaxy workflow screenshot.

\section{Conclusion}
\label{sec:conclusion}

Bioinformatics workflow construction exposes a core limitation of long-horizon LLM agents: without persistent process memory, the agent repeatedly spends context on software recovery rather than scientific reasoning. We introduced Process-Reward Tactic Evolution, a Galaxy-based training framework that converts real workflow rollouts into reusable tactics for workflow authoring, schema matching, invocation, repair, artifact collection, and biological checking. On BioWorkflow Bench and BioAgent Bench, this tactic state improves long and xlong task performance, biological correctness, and execution efficiency. The broader conclusion is that scientific agents need verifier-backed, reusable process memory when they operate in complex software environments.

\section*{Limitations}

Our experiments are centered on Galaxy, so performance depends on public-server availability, queue behavior, installed tool versions, and dataset accessibility. Some workflows are too backend-specific or long-running for perfectly stable shared-infrastructure evaluation. Biological assertions are useful but partial; passing them does not replace expert review of downstream scientific interpretation. The tactic updater must also remain conservative, because an incorrect reusable repair can harm future tasks.

\section*{Ethics Statement}

The benchmark tasks, workflow templates, documentation, and input data used in this paper are derived from public bioinformatics resources. The system is intended for workflow-construction research and should not be used as a substitute for expert biological or clinical judgment. We report converted estimates separately from actual \texttt{test35} runs where applicable. After publication, we plan to release the code, task manifests, prompts, evaluator configuration, and generated tactic artifacts needed to reproduce the experiments, subject to public-service access limits and any required cleanup of authentication material.

\clearpage
\onecolumn
\appendix
\section{Related Work}
\label{app:related-work}

\subsection{Long-Horizon Science and Bioinformatics Agents}

BixBench, ScienceAgentBench, and SciAgentGym move science-agent evaluation beyond static question answering toward executable, multi-step tasks \citep{mitchener2025bixbench,chen2025scienceagentbench,shen2026sciagentgym}. BioAgent Bench narrows this direction to bioinformatics pipelines with executable traces and perturbation tests \citep{fa2026bioagentbench}. BioMaster, GeneAgent, and ReflecTool further show that biomedical agents need planning, retrieval, tool use, and self-verification \citep{su2025biomaster,geneagent2025,liao2024reflectool}. BioWorkflow is complementary: it centers long-horizon Galaxy workflow-DAG execution and asks whether agents can learn reusable process tactics from repeated verifiable interactions.

\subsection{Bioinformatics Workflow Automation}

Workflow-specific work studies how LLMs retrieve, generate, or convert scientific workflows. Prompt-to-Pipeline evaluates LLM-generated Galaxy and Nextflow workflows, Galaxy retrieval work ranks reusable workflows for a task, and Snakemaker converts ad hoc analyses into Snakemake workflows \citep{alam2025prompttopipeline,cynthia2025galaxyretrieval,masera2025snakemaker}. WfBench and community workflow resources such as IWC motivate executable workflow artifacts and benchmarks \citep{coleman2022wfbench,iwc}. These systems address workflow access and representation; our agent must also bind inputs, invoke workflows, monitor jobs, inspect artifacts, repair failures, and update tactics from traces.

\subsection{Memory, Skills, and Weight-Free Agent Learning}

BioWorkflow also belongs to externalized agent learning. Reflexion and ExpeL store verbal reflections or distilled insights \citep{shinn2023reflexion,zhao2024expel}; Voyager and Skill-Pro learn reusable skills or procedures \citep{wang2023voyager,mi2026skillpro}; Agent Workflow Memory, A-MEM, and ReasoningBank organize workflow, graph, or reasoning memories \citep{wang2024awm,xu2025amem,ouyang2026reasoningbank}; and Memento, Training-Free GRPO, GEPA, and EvoTest adapt memory, prompt, or agent-system state across trials \citep{zhou2025memento,trainingfreegrpo2025,agrawal2026gepa,he2026evotest}. Graph of Skills and SkillNet emphasize that growing skill libraries require structure, evaluation, and dependency-aware retrieval \citep{liu2026gos,liang2026skillnet}. Our tactic library specializes this direction to scientific workflow software: each tactic is a typed, executable, verifier-backed process unit with applicability conditions, software actions, postconditions, failure signatures, and repair routes grounded in Galaxy evidence.

\subsection{Process Rewards, Verifiers, and Harnesses}

Process-reward and verifier-harness work shows why intermediate evidence matters. Process Reward Agents use step-level rewards, Meta-Harness optimizes harness behavior from source code, scores, and traces, and DoVer validates debugging hypotheses through intervention and replay \citep{sohn2026pra,lee2026metaharness,ming2025dover}. Agentic Aggregation and verifier-design work similarly treat parallel trajectories and checkers as inspectable evidence for long-horizon agents \citep{lee2026aggagent,rosset2026verifiers}. BioWorkflow transfers this principle to bioinformatics workflows, where process evidence comes from BioBlend responses, gxformat2 graph checks, datatype and collection metadata, Galaxy job states, output artifacts, and biological assertions.

\section{BioWorkflow Bench Data Preparation}
\label{app:bench-preparation}

We built BioWorkflow Bench to study executable workflow construction rather than short-answer biomedical reasoning. We first surveyed existing bioinformatics-agent and workflow resources, including BixBench, BioAgent Bench, CWL, WDL, Nextflow, and Galaxy. BixBench emphasizes coding and multiple-choice reasoning, which does not directly match our target of workflow graph construction and live execution. BioAgent Bench provides useful executable bioinformatics tasks and a local sandbox, but the public set contains only ten relatively short tasks. It is therefore valuable as an external comparison but insufficient as the main benchmark for long-horizon workflow learning.

Galaxy provides the strongest basis for our setting. It has a public execution backend through usegalaxy, maintained tool wrappers, BioBlend and gxformat2 APIs, and large public workflow resources. The IWC workflows are especially useful because they contain peer-reviewed Galaxy workflow graphs and reproducibility metadata. GTN materials provide additional training-style workflow examples. These resources allow the agent to operate against a real platform without requiring us to install every bioinformatics tool and database locally.

We chose Galaxy over CWL, Nextflow, and WDL for this benchmark because Galaxy combines public tasks, workflow graphs, assertions, documentation, and a free shared execution backend. Container-centered workflow systems are important future targets, but they would substantially increase evaluation complexity because each task may require local container orchestration, reference data, and environment-specific execution logic. Our current design therefore focuses on Galaxy while keeping the task representation compatible with future workflow-platform expansion.

Tables~\ref{tab:appendix-training-tasks} and~\ref{tab:appendix-test-tasks} list the 26 training tasks and 35 held-out test tasks. In these tables, \textit{Tools} counts unique Galaxy workflow tools, \textit{Depth} is the longest dependency-chain length in the workflow graph, \textit{Inputs} counts task inputs, and \textit{Asrts} counts available assertion checks. BioAgent Bench does not expose the same gold Galaxy workflow DAG metadata as BioWorkflow Bench; in the main text, its tool and depth statistics therefore use the adopted Galaxy-run tool counts and evaluator step-count proxy. Table~\ref{tab:appendix-galaxy-docs} lists the initial Galaxy documentation projects exposed to the agents.

\clearpage
\begin{table}[p]
\centering
\tiny
\setlength{\tabcolsep}{2.4pt}
\renewcommand{\arraystretch}{0.74}
\begin{tabularx}{\linewidth}{@{}llYrrrr@{}}
\toprule
Src & Size & Topic & Tools & Depth & Inputs & Asrts \\
\midrule
IWC & small & SRA accession download & 3 & 4 & 1 & 3 \\
IWC & small & Raw-read quality control & 7 & 6 & 2 & 0 \\
IWC & small & 10X single-cell FASTQ to matrix & 3 & 4 & 4 & 4 \\
IWC & small & ChIP-seq single-end processing & 7 & 6 & 1 & 6 \\
IWC & small & Flye genome assembly & 4 & 3 & 1 & 10 \\
GTN & small & Braker3 genome annotation & 5 & 4 & 3 & 0 \\
GTN & medium & NGS variant-analysis tutorial & 13 & 12 & 2 & 45 \\
IWC & medium & RNA-seq single-end quantification & 8 & 6 & 2 & 9 \\
IWC & medium & RNA-seq differential expression & 13 & 9 & 5 & 0 \\
GTN & medium & Pseudobulk differential expression & 10 & 9 & 1 & 0 \\
IWC & medium & CUT\&RUN/CUT\&TAG analysis & 9 & 8 & 2 & 8 \\
GTN & medium & Hi-C analysis with HiCExplorer & 11 & 5 & 9 & 2 \\
IWC & medium & Generic WGS paired-end variants & 12 & 10 & 3 & 0 \\
IWC & medium & Assembly decontamination & 9 & 10 & 1 & 0 \\
IWC & medium & Bacterial genome annotation & 7 & 5 & 1 & 11 \\
IWC & medium & DADA2 paired-end amplicons & 11 & 10 & 10 & 0 \\
GTN & medium & QCxMS EI+ mass-spectra prediction & 11 & 13 & 1 & 0 \\
IWC & medium & HyPhy CAPHEINE compare & 10 & 13 & 6 & 0 \\
IWC & long & ATAC-seq accessibility profiling & 21 & 14 & 2 & 11 \\
IWC & long & SARS-CoV-2 ARTIC PE variation & 20 & 17 & 5 & 0 \\
IWC & long & GROMACS dcTMD calculations & 13 & 10 & 2 & 0 \\
GTN & xlong & Differential isoform expression & 35 & 23 & 18 & 5 \\
IWC & xlong & Scanpy single-cell clustering & 14 & 26 & 3 & 1 \\
IWC & xlong & VGP3 HiFi genome assembly & 21 & 11 & 4 & 11 \\
IWC & xlong & VGP6 duplicate-contig purging & 16 & 18 & 6 & 13 \\
IWC & xlong & Metagenome-assembled genomes & 29 & 20 & 4 & 0 \\
\bottomrule
\end{tabularx}
\caption{Training tasks used by the BioWorkflow Bench curriculum. Tools, Depth, Inputs, and Asrts count workflow tools, longest dependency-chain depth, input datasets, and output assertions.}
\label{tab:appendix-training-tasks}

\vspace{0.55em}
\begin{tabularx}{\linewidth}{@{}llYrrrr@{}}
\toprule
Src & Size & Topic & Tools & Depth & Inputs & Asrts \\
\midrule
IWC & small & Short-read QC and trimming & 2 & 3 & 2 & 6 \\
IWC & small & Bacterial genome assembly & 5 & 4 & 2 & 5 \\
GTN & small & Single-cell RNA-seq QC & 4 & 4 & 3 & 4 \\
IWC & small & BigWig replicate averaging & 2 & 3 & 4 & 2 \\
IWC & small & HyPhy comparative selection & 4 & 5 & 5 & 4 \\
IWC & medium & OpenMS MetaProSIP proteomics & 8 & 7 & 2 & 0 \\
IWC & medium & MAKER genome annotation & 7 & 4 & 5 & 0 \\
IWC & medium & Post-assembly bacterial QC & 8 & 6 & 1 & 16 \\
IWC & medium & Fragment docking and scoring & 10 & 8 & 4 & 0 \\
IWC & medium & Haploid WGS paired-end variants & 13 & 11 & 6 & 1 \\
IWC & medium & Long-read assembly polishing & 2 & 9 & 2 & 0 \\
IWC & medium & Single-cell pseudobulk edgeR & 10 & 11 & 1 & 0 \\
IWC & medium & VGP1 HiFi k-mer profiling & 10 & 9 & 5 & 0 \\
GTN & medium & Metagenomic sequencing assembly & 9 & 5 & 12 & 20 \\
GTN & medium & Tuberculosis variant analysis & 14 & 6 & 5 & 8 \\
IWC & long & ATAC/CUT\&RUN consensus peaks & 18 & 15 & 3 & 7 \\
IWC & long & Viral paired-end consensus & 22 & 12 & 6 & 0 \\
GTN & long & RNA-seq QC, mapping, counting & 9 & 6 & 6 & 34 \\
GTN & long & Machine-learning clustering & 3 & 4 & 3 & 9 \\
GTN & long & Somatic variant and CNV discovery & 16 & 16 & 5 & 0 \\
IWC & long & ChIP-seq paired-end consensus peaks & 17 & 14 & 3 & 7 \\
IWC & long & ChIP-seq single-end consensus peaks & 17 & 14 & 3 & 7 \\
IWC & long & MAG taxonomy annotation & 12 & 19 & 4 & 0 \\
IWC & long & SARS-CoV-2 ONT ARTIC variation & 21 & 13 & 3 & 0 \\
IWC & long & SARS-CoV-2 variation reporting & 14 & 17 & 5 & 2 \\
IWC & xlong & Metagenomic genes catalogue & 24 & 18 & 2 & 13 \\
IWC & xlong & Hi-C contact map for curation & 20 & 17 & 5 & 19 \\
GTN & xlong & Tn-seq essential-gene detection & 12 & 21 & 5 & 3 \\
IWC & xlong & MAG binning evaluation & 26 & 14 & 4 & 0 \\
IWC & xlong & Influenza consensus and subtyping & 28 & 32 & 6 & 0 \\
IWC & xlong & Generic variation reporting & 12 & 19 & 1 & 0 \\
IWC & xlong & VGP5 HiFi trio-phased assembly & 21 & 11 & 10 & 8 \\
IWC & xlong & VGP6b haplotype duplicate purging & 15 & 13 & 6 & 10 \\
GTN & xlong & Metatranscriptomics & 22 & 9 & 2 & 6 \\
GTN & xlong & Single-cell filtering and exploration & 18 & 18 & 1 & 4 \\
\bottomrule
\end{tabularx}
\caption{Test35 tasks used for held-out BioWorkflow Bench evaluation. Tools, Depth, Inputs, and Asrts follow the same definitions as Table~\ref{tab:appendix-training-tasks}.}
\label{tab:appendix-test-tasks}
\end{table}

\clearpage
\begin{table}[t!]
\centering
\small
\setlength{\tabcolsep}{4.0pt}
\renewcommand{\arraystretch}{1.08}
\begin{tabularx}{\linewidth}{@{}lYY@{}}
\toprule
Doc Project & Main Coverage & Repository \\
\midrule
Galaxy Hub / UseGalaxy & Platform operation, histories, datasets, quotas, sharing, and API overview. & \href{https://github.com/galaxyproject/galaxy-hub}{\texttt{galaxy-hub}} \\
Galaxy Core & REST objects, tool schema, jobs, workflows, invocations, datatypes, and provenance. & \href{https://github.com/galaxyproject/galaxy}{\texttt{galaxy}} \\
BioBlend & Python API for histories, datasets, tools, jobs, workflows, import, invoke, and monitor. & \href{https://github.com/galaxyproject/bioblend}{\texttt{bioblend}} \\
gxformat2 & YAML workflow authoring, graph validation, and conversion to Galaxy \texttt{.ga}. & \href{https://github.com/galaxyproject/gxformat2}{\texttt{gxformat2}} \\
Planemo & Workflow tests, test YAML, assertions, invocation tracking, and output checks. & \href{https://github.com/galaxyproject/planemo}{\texttt{planemo}} \\
IWC & Expert Galaxy workflow packages, gold graphs, tests, metadata, and reproducibility conventions. & \href{https://github.com/galaxyproject/iwc}{\texttt{iwc}} \\
GTN & Galaxy tutorial workflows, data links, topic material, examples, and biological explanations. & \href{https://github.com/galaxyproject/training-material}{\texttt{training-material}} \\
\bottomrule
\end{tabularx}
\caption{P0 Galaxy documentation sources used for workflow execution, authoring, testing, and benchmark construction.}
\label{tab:appendix-galaxy-docs}
\end{table}

\section{Prompts and Tactics Evolution}
\label{app:experimental-comparisons}

This appendix expands the training-time prompt structure and the evolution of the tactic library. The purpose is to make the method reproducible at the level of agent contracts: what the solver is allowed to see, what the reviewer can inspect, and how the updater converts public process evidence into reusable tactic files. The three prompts are not independent chat templates. Together they form a pipeline: the solver produces a Galaxy execution trajectory, the reviewer separates private evaluation analysis from public training feedback, and the tactic updater writes gated library patches from the public evidence.

Figure~\ref{fig:solver-inference-prompt} shows the solver/inference prompt. Its most important role is to define the visibility boundary and the required operating loop. The solver is instructed to work only inside the run directory, use the configured Galaxy backend, read the sandbox curriculum and allowed documentation, discover live tool schemas through BioBlend, construct gxformat2 workflows, run local checks, import and invoke workflows, and package provenance. The prompt deliberately does not expose gold workflows, exact rubrics, hidden assertions, or benchmark repository paths.

\begin{figure}[t!]
\centering
\vspace*{-2.0cm}
\setlength{\fboxsep}{3pt}
\fcolorbox{gray!55}{gray!5}{%
\begin{minipage}{0.94\linewidth}
\colorbox{gray!30}{\makebox[\linewidth][c]{\strut\textbf{solver/inference prompt}}}
\vspace{0.45em}

{\footnotesize
\setlength{\parskip}{0.42em}
\raggedright
You are the BioWorkflow Galaxy agent.

You run locally through Codex CLI using model gpt-5.5 with xhigh reasoning. Your execution backend is \texttt{\{galaxy\_url\}}. Your control surface is Python, the BioBlend API client, gxformat2, Planemo, and a Jupyter notebook.

\textbf{Hard constraints:}
- You may only use files under this run directory and the explicit paths named in \texttt{\{policy\}}.
- Do not read any gold \texttt{.ga}, exact rubric, or benchmark tests.
- Do not access IWC, GTN, task repositories, answer repositories, search engines, or dataset pages not explicitly listed in \texttt{data\_manifest.json}.
- Treat \texttt{data\_manifest.json} as the authoritative task input list.
- Data transport policy: when a task input has an explicit \texttt{download\_url}, prefer UseGalaxy server-side URL ingestion into the task history...
- Do not print or save the Galaxy API key. Load it from environment variables only.
- \texttt{GALAXY\_URL} is set to \texttt{\{galaxy\_url\}}. Use that configured backend for live schema discovery, data staging, workflow import, invocation, and job polling.
- \texttt{\{network\_instructions\}}

\textbf{Available helper commands:}
\texttt{notebook\_cli init \{workspace/agent.ipynb\}}; \texttt{notebook\_cli append ... --kind code --file cell.py}; \texttt{notebook\_cli run-cell ... CELL\_INDEX --kernel bioworkflow-galaxy --cwd \{workspace\}}; ...

\textbf{Your required workflow:}
1. Create and maintain \texttt{\{workspace/agent.ipynb\}}. Prefer \texttt{run-cell}; make Galaxy cells idempotent; treat \texttt{output/run\_summary.json} and \texttt{results/run\_summary.json} as canonical live-state pointers.
2. Inspect \texttt{README.md}, \texttt{catalog.md}, and relevant \texttt{CURRICULUM\_INDEX.md} files before live UseGalaxy submission.
3. Mine sandbox documentation for Galaxy, BioBlend, gxformat2, Planemo, notebook, and UseGalaxy lifecycle procedures.
4. Use BioBlend to verify the Galaxy connection and discover relevant tools/data interfaces.
5. Inspect tool schemas with \texttt{gi.tools.show\_tool(..., io\_details=True)}.
6. Record tool IDs, input names, output names, and graph assumptions before workflow import.
7. Build the DAG workflow by discovery; run local checks; submit only after checks pass.
8. Debug with import errors, invocation/job states, dataset states, and downloaded small logs.
9. Save provenance, assertion trace, and final deliverables under the output/results directories.

\textbf{\#\# Shared sandbox curriculum}
Use the curriculum for BioBlend API usage, gxformat2 templates, Planemo/local check interpretation, UseGalaxy lifecycle, Galaxy tool schema discovery, notebook operations, and provenance/report packaging. Treat visibility labels as policy...

\textbf{\#\# Tactics-training discovery rule}
Infer candidate tool families from the biological task statement; search only allowed sandbox docs and live configured-Galaxy tool catalog; inspect exact tool schemas before wiring a DAG; use notebook experiments, local checks, Galaxy feedback, and trajectory writing as environment signals.

\textbf{\#\# Task description} \texttt{\{task\}}

\textbf{\#\# Data manifest} \texttt{\{data\_manifest\}}

\textbf{\#\# Expected result schema} Schema of output types and qualitative checks, not hidden gold.

\textbf{\#\# Tactics library}
Use these tactics when deciding how to write Galaxy API code, construct gxformat2, debug Galaxy errors, and manage the UseGalaxy lifecycle. \texttt{\{tactics\}} \texttt{\{controller\_hint\_section\}}
\par}
\end{minipage}}
\caption{Source-faithful outline of the solver/inference prompt.}
\label{fig:solver-inference-prompt}
\end{figure}
\clearpage

The reviewer prompt in Figure~\ref{fig:private-reviewer-prompt} is broader because the reviewer is part of the training infrastructure, not the solver. It may inspect private evaluator artifacts and gold comparisons in order to diagnose why a rollout succeeded or failed. However, it must write two outputs: a private report for analysis and a sanitized public feedback draft for training. This split is important because the tactic updater must not receive hidden gold details; it should learn general process lessons such as tool-schema mismatch, collection wiring, job-state handling, or artifact packaging failures.

\begin{figure}[t!]
\centering
\vspace*{-0.8cm}
\setlength{\fboxsep}{3pt}
\fcolorbox{gray!55}{gray!5}{%
\begin{minipage}{0.94\linewidth}
\colorbox{gray!30}{\makebox[\linewidth][c]{\strut\textbf{reviewer prompt}}}
\vspace{0.5em}

{\small
\setlength{\parskip}{0.45em}
\raggedright
You are the BioWorkflow TaskLevelLLMReviewer.

Review every rollout for one benchmark task. You are a private evaluator and may inspect the run directories, private evaluator directories, gold workflows, gold tests, rubrics, programmatic evaluation features, notebooks, Codex event logs, tool-schema probes, local checks, Galaxy API traces, and result files.

\textbf{Task metadata:}
\texttt{\{json.dumps(task.to\_json(), indent=2)\}}

\textbf{Rollout records:}
\texttt{\{json.dumps(records, indent=2)\}}

\textbf{Reviewer operating protocol:}
- Start by reading \texttt{\{context\_manifest\}}. It is a curated file manifest for rollout and private-evaluation artifacts.
- Do not run broad discovery commands such as \texttt{find}, \texttt{rg --files}, \texttt{ls -R}, or recursive directory walks.
- Use targeted reads against paths from the manifest: \texttt{sed -n}, \texttt{head}, \texttt{tail}, \texttt{python -m json.tool}, or small Python snippets that open explicitly listed files.
- If a file is omitted but genuinely needed, inspect only one named subdirectory from one rollout with a hard cap of 50 path entries.
- Prefer \texttt{results/run\_summary.json}, \texttt{output/run\_summary.json}, \texttt{results/assertion\_trace.json}, private evaluator JSON/YAML, and the final Codex message before large event logs or notebooks.

\textbf{Write two files:}

1. \textbf{Private report:} \texttt{\{private\_report\}}.
This may mention private gold details, exact graph differences, assertion details, missing gold branches, and rollout-specific private comparisons. Compare rollouts against each other and against the gold workflow/tests. Separate process-quality failures from final-answer failures.

2. \textbf{Public sanitized feedback:} \texttt{\{public\_draft\}}.
This will be gate-checked and may be shown to the tactics updater. Do not include exact gold step IDs, tool versions, assertions, private file paths, output values, test snippets, repository paths, task item IDs, source URLs, checksums, or case-identifying strings. Name the controller mode exactly from the run summaries... Include task-family-level and process-level lessons only when supported by observed trajectories.

Do not modify solver run files, private eval files, or tactics checkpoints.
\par}
\end{minipage}}
\caption{Source-faithful outline of the reviewer prompt.}
\label{fig:private-reviewer-prompt}
\end{figure}
\clearpage

Figure~\ref{fig:tactic-updater-prompt} shows the tactic updater prompt. The updater treats the tactic checkpoint as a maintained linked library rather than an append-only reflection log. It may add, update, delete, rename, merge, or deprecate Markdown tactics, but every active tactic must include activation conditions, procedures, termination checks, failure modes, verifier expectations, related links, and public evidence. This is where the method differs from prompt-only evolution: the update target is a structured process memory, not one global instruction.

\begin{figure}[t!]
\centering
\vspace*{-0.8cm}
\setlength{\fboxsep}{3pt}
\fcolorbox{gray!55}{gray!5}{%
\begin{minipage}{0.94\linewidth}
\colorbox{gray!30}{\makebox[\linewidth][c]{\strut\textbf{tactic updater prompt}}}
\vspace{0.5em}

{\footnotesize
\setlength{\parskip}{0.42em}
\raggedright
You are the BioWorkflow TacticUpdaterAgent.

Your job is to propose a linked-tactics patch bundle from public training signals only. A tactics checkpoint is a skill library: multiple focused tactic files connected by an index and cross-links. It must not collapse a whole stage into one omnibus reflection file.

\textbf{Use this patch shape:}
- Treat the tactics checkpoint as a maintained linked skill library, not an append-only reflection log.
- Decide the library edit yourself from public evidence and current tactics: add, update, delete, rename, merge, deprecate, reorganize, or leave unchanged.
- Do not target a fixed number of files.
- Keep \texttt{INDEX.md} and topic indexes consistent when retrieval cues, links, file paths, or load order change.
- Prefer topic directories: \texttt{api/}; \texttt{workflow\_authoring/}; \texttt{execution/}; \texttt{repair/}; \texttt{verification/}; \texttt{task\_families/}; ...

\textbf{You may read:}
- current tactics: \texttt{\{current\_tactics\}}
- this public group summary embedded below
- public trajectory summaries under \texttt{\{proposal\_dir/public\_trajectories\}}
- public task-level LLM review summaries under \texttt{\{proposal\_dir/public\_task\_reviews\}}

You must not use private evaluator files, gold workflows, gold tests, original benchmark repository paths, search engines, or task-specific hidden data.

\textbf{Write proposed tactic files under:}
\texttt{\{proposal\_dir/files\}}

Use repository-relative paths such as \texttt{repair/usegalaxy\_collection\_wiring.md}, \texttt{workflow\_authoring/gxformat2\_terminal\_outputs.md}, \texttt{verification/reference\_artifact\_contract.md}, \texttt{task\_families/rnaseq\_velocity.md}, or \texttt{INDEX.md}. Only Markdown files are allowed.

Also write \texttt{\{proposal\_dir/proposal\_manifest.json\}}, with changes containing \texttt{path}, \texttt{action}, \texttt{new\_path}, \texttt{rationale}, and public \texttt{evidence}.

Each active tactic file must include activation, procedure, termination, failure modes, verifier expectations, links to related tactics, and evidence.

Do not write case-specific rescue hints. Convert observed failures into general, reusable procedures. Extract a semantic group advantage for BioWorkflow tactics.

\textbf{Training unit:} \texttt{\{task\_id\}}

Compare rollouts by process reward, not only final score. Identify which workflow-construction decisions, API usage patterns, repair actions, and lifecycle choices caused relative success or failure. Propose only linked tactic deltas with activation, procedure, termination, verifier evidence, and links to existing tactics.

\textbf{Contexts:} \texttt{\{rollout\_summaries\}}, \texttt{\{group\_summary\}}, \texttt{\{trajectory\_context\}}, \texttt{\{task\_review\_context\}}.
\par}
\end{minipage}}
\caption{Source-faithful outline of the tactic updater prompt.}
\label{fig:tactic-updater-prompt}
\end{figure}
\clearpage

Table~\ref{tab:tactics-evolution} summarizes the resulting library growth. The seed checkpoint contains a small set of general API, workflow-authoring, lifecycle, and verification tactics. After curriculum training, the process-reward setting expands into repair tactics and task-family contracts, while outcome-only training grows more slowly and remains concentrated on generic completion and biology triage. This table is therefore the quantitative counterpart to the prompt pipeline above: process feedback changes what kind of memory is written.

\begin{table}[t!]
\centering
\footnotesize
\setlength{\tabcolsep}{3.4pt}
\renewcommand{\arraystretch}{1.08}
\begin{tabularx}{\linewidth}{@{}llrrY@{}}
\toprule
Checkpoint & Complexity & Files & Tokens & Main lesson \\
\midrule
seed & n/a & 8 & 6,669 & Initial generic tactics for BioBlend access, gxformat2 authoring, lifecycle discipline, and verification. \\
batch1 & Small 4; medium 2 & 13 & 12,412 & Early live failures become cross-cutting graph, collection, terminal-repair, timeout, and chromatin contracts. \\
batch2 & Medium 6 & 18 & 19,070 & Adds terminal semantic checks and medium-task family contracts for RNA-seq, pseudobulk, decontamination, and variant/report workflows. \\
batch3 & Medium 2; long 2; xlong 2 & 23 & 27,102 & Generalizes from long/xlong tasks into mapped scalar validation, chemistry, DE, molecular dynamics, MAG, assembly QC, and ARTIC/variant updates. \\
batch4 & Medium 2; long 1; xlong 3 & 28 & 33,700 & Adds domain contracts for the remaining difficult families: Hi-C, HyPhy, isoform switching, Scanpy clustering, purge-dups, and ATAC-specific chromatin details. \\
\bottomrule
\end{tabularx}
\caption{Tactics Evolution. Files counts Markdown files, Tokens is the approximate tactic-library token count, and Complexity follows the curriculum progression, with Tiny normalized to Small.}
\label{tab:tactics-evolution}
\end{table}

\section{Case Study: XLong Hi-C Workflow}
\label{app:case-study}

This appendix expands the xlong Hi-C case discussed in Section~\ref{sec:results}. The goal is not to add another leaderboard result, but to inspect how the learned tactics affect a live Galaxy trajectory. The case is useful because it requires long-horizon software interaction: the agent must reason about assembly curation, Hi-C alignment, contact-map generation, Galaxy tool schemas, workflow import and invocation, failure repair, and final artifact collection.

\subsection{Case Selection and Evidence}

We analyze the held-out Hi-C contact-map task for assembly manual curation. The task is xlong, belongs to the same \texttt{test35} split used in the main evaluation, and exposes the main failure modes of workflow agents: missing tool wrappers, datatype and collection wiring, invocation failures, optional diagnostic branches, and output-family accounting. The representative run uses the batch-4 tactic checkpoint. Its execution trace contains policy checks, tactic retrieval, notebook cells, live Galaxy API calls, workflow draft creation, graph upload, invocation, repair attempts, and final evidence packaging.

The visibility boundary remains the same as in training. The solver sees only the task bundle, public documentation, live Galaxy responses, and learned tactics. It does not see hidden gold workflows, exact private tests, or benchmark answer paths. This makes the case a process analysis of tactic use rather than evidence of gold-workflow access.

\subsection{Tactics Aligned to Execution Turns}

\begin{figure}[p]
\centering
\vspace*{-1.0cm}
\definecolor{tacticBlue}{RGB}{225,238,250}
\definecolor{traceGreen}{RGB}{228,242,232}
\definecolor{linkGray}{RGB}{120,120,120}
\begin{tikzpicture}[
  font=\small,
  tactic/.style={draw=blue!35, fill=tacticBlue, rounded corners=2pt, align=left, text width=6.30cm, inner sep=4.0pt},
  trace/.style={draw=green!35!black, fill=traceGreen, rounded corners=2pt, align=left, text width=9.25cm, inner sep=4.0pt},
  arrow/.style={-{Latex[length=1.5mm]}, draw=linkGray, line width=0.35pt},
  rowarrow/.style={-{Latex[length=1.2mm]}, draw=gray!45, line width=0.28pt}
]
\node[font=\normalsize\bfseries] at (0,1.05) {Activated tactics};
\node[font=\normalsize\bfseries] at (8.45,1.05) {Agent execution trace};

\foreach [count=\i] \y/\l/\r in {
0/{Lifecycle hygiene; policy discipline}/{Records allowed docs and restricted sources before any Galaxy call; avoids gold workflow access and writes the doc-mining report.},
-1.60/{Tactic selection; Hi-C and assembly contracts}/{Loads Hi-C, assembly QC, semantic-graph, live-schema, and local-check tactics before planning.},
-3.20/{BioBlend bootstrap; notebook state}/{Creates an idempotent notebook, output directories, run summaries, and a public role contract.},
-4.80/{Live schema and artifact evidence}/{Queries live UseGalaxy schemas; when narrow search misses tools, switches to a full live tool catalog and filters locally.},
-6.40/{Role contract before graph writing}/{Maps task prose to obligations: haplotypes, gap/telomere tracks, Hi-C statistics, Pretext maps, and coverage outputs.},
-8.00/{gxformat2 local check loop}/{Writes gxformat2, converts to native workflow, lints, and patches output declarations from live schemas before import.},
-9.60/{Collection wiring; invocation hygiene}/{Stages server-side URLs, creates Hi-C and PacBio collections, imports the workflow, and invokes it in one history.},
-11.20/{Terminal failure repair gate}/{Stops after terminal failure; preserves failed-job evidence from the legacy BWA-MEM wrapper instead of blind retrying.},
-12.80/{Wrapper-specific bounded repair}/{Repairs BWA-MEM to minimap2, then BWA-MEM2, then removes optional PretextSnapshot after schema-specific failure.},
-14.40/{Semantic smoke audit; final packaging}/{Final repair completes 22 Galaxy jobs and 17 workflow outputs; downloads artifacts, writes assertion trace, and checks for leakage.}
} {
  \node[tactic] (L\i) at (0,\y) {\l};
  \node[trace] (R\i) at (8.45,\y) {\r};
  \draw[arrow] (L\i.east) -- (R\i.west);
}

\foreach \i/\j in {1/2,2/3,3/4,4/5,5/6,6/7,7/8,8/9,9/10} {
  \draw[rowarrow] (L\i.south) -- (L\j.north);
  \draw[rowarrow] (R\i.south) -- (R\j.north);
}
\end{tikzpicture}
\caption{Tactics-use timeline for the representative xlong Hi-C task. The left lane shows activated process-memory tactics; the right lane shows aligned execution steps and concrete Galaxy/workflow evidence.}
\label{fig:tactics-use-hic-trace}
\end{figure}
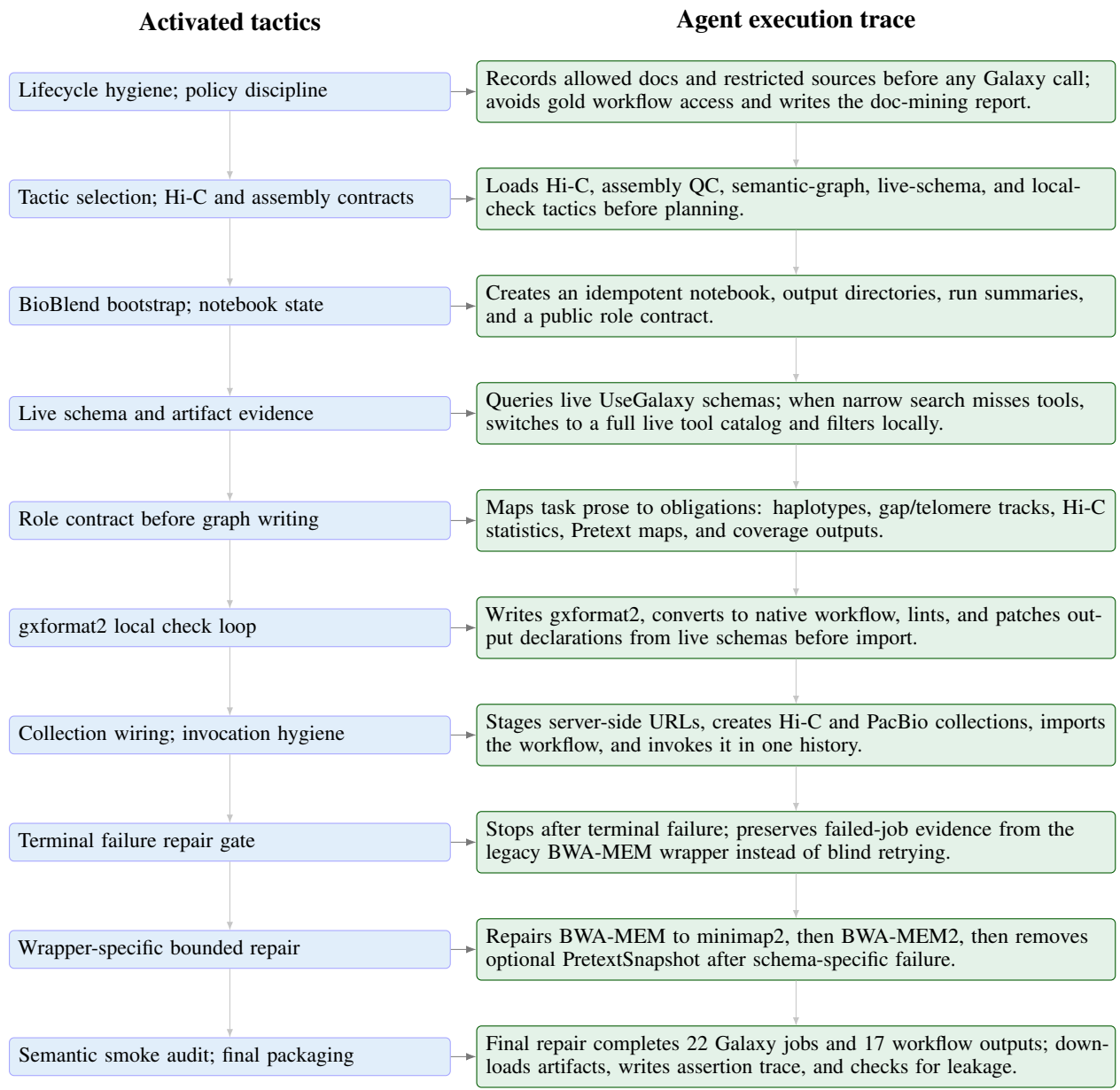

Figure~\ref{fig:tactics-use-hic-trace} aligns selected tactic elements with execution trace milestones. The left side contains tactic-level guidance such as reading visibility policy, deriving output-family obligations, using live Galaxy schema discovery, checking gxformat2 before import, and preserving repair provenance. The right side shows the corresponding trajectory events. The important pattern is that the agent does not simply reason longer. It follows a more direct operational route: establish a role contract, inspect schemas before wiring tools, upload a draft graph, use Galaxy failure state as typed feedback, and apply bounded repairs.

\clearpage

The first major transition is from biological prose to a workflow role contract. The agent identifies curation assembly construction, assembly statistics, gap and telomere tracks, Hi-C raw and deduplicated alignments, Pretext contact maps, and coverage outputs before writing the executable graph. This is the intended use of tactics: they provide reusable verifier expectations and software-operation procedures, not stored answers for a specific benchmark case.

\subsection{Repair Behavior}

The trace also illustrates why process rewards are more useful than final scores alone. A failed Galaxy invocation is not merely a negative outcome. It contains typed evidence: which job failed, which wrapper emitted the error, which input or parameter was implicated, and which downstream branch should be repaired. In this case, the agent uses Galaxy state to stop polling, inspect the failed job, check live schemas, patch only the implicated branch, rerun local workflow checks, and re-invoke. Several baseline agents consume many more tokens without reaching the same repair depth because they do not convert platform feedback into bounded workflow edits.

\subsection{Constructed Galaxy Workflow}

\begin{figure}[p]
\centering
\includegraphics[width=\linewidth]{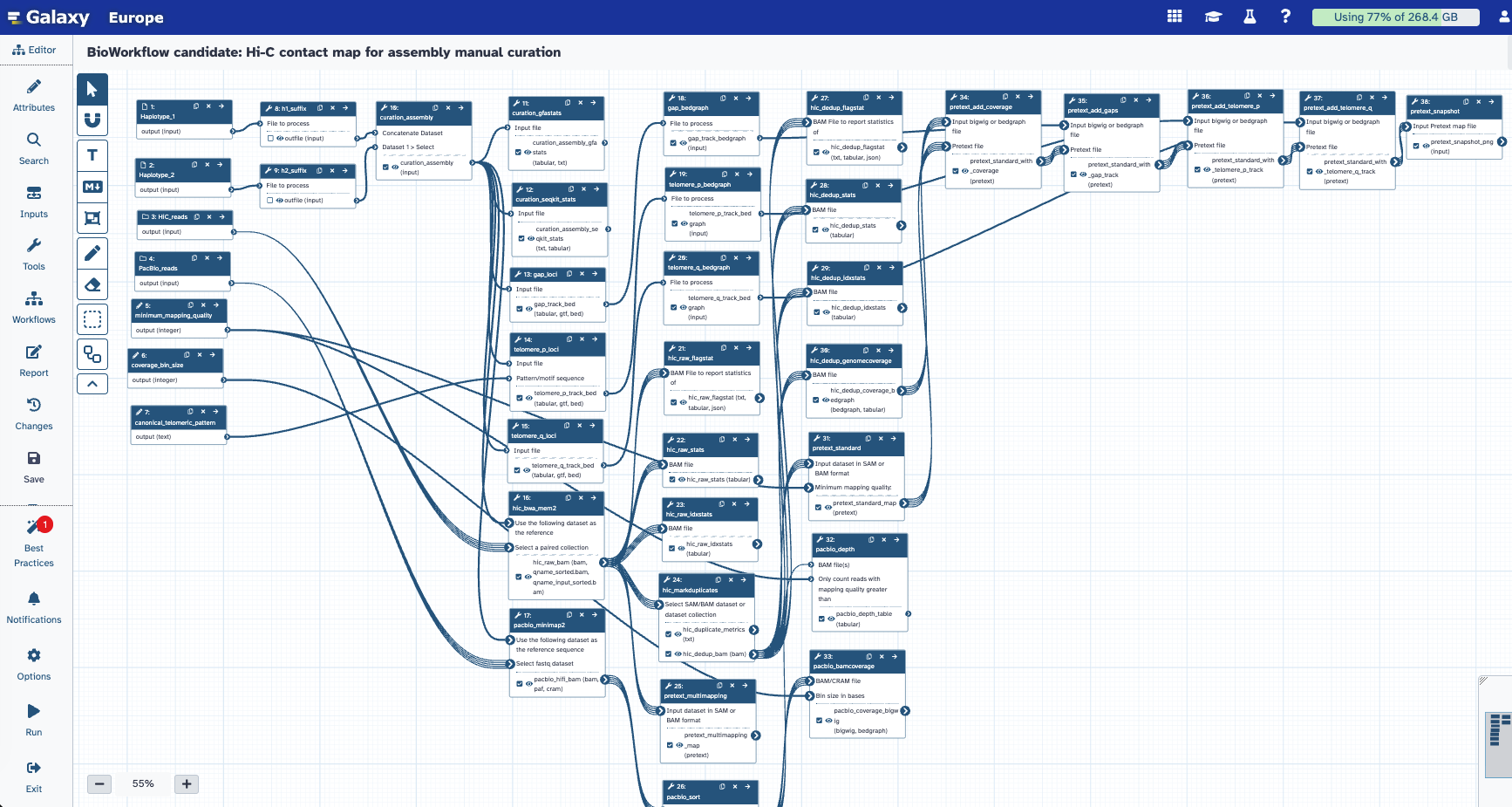}
\caption{Representative xlong Galaxy workflow constructed by the post-training PRTE Agent. The workflow is the final Hi-C contact-map workflow for assembly manual curation, captured from the usegalaxy.eu website.}
\label{fig:xlong-workflow-screenshot}
\end{figure}

The final workflow screenshot in Figure~\ref{fig:xlong-workflow-screenshot} is included as a concrete example of the workflow graph that the agent constructs and executes on the usegalaxy backend. The graph is not a perfect reproduction of the public expert workflow; it is a compact executable reconstruction produced under the solver visibility boundary. It stages haplotype FASTA inputs, Hi-C paired-read collections, and long-read inputs, then combines assembly preparation, statistics, alignment, contact-map generation, and coverage-output branches.

This case should therefore be interpreted at two levels. At the mechanism level, learned tactics make the trajectory more auditable and repairable: the agent reads relevant procedures, uses live schema evidence, uploads an executable graph, diagnoses errors, and keeps final outputs separate from superseded failed attempts. At the benchmark level, xlong workflow recovery remains difficult. The remaining errors show that agents still need stronger branch-level topology recovery and optional-output reasoning, especially when expert workflows contain diagnostic or reporting branches that are not necessary for terminal execution but matter for exact biological matching.


\begin{thebibliography}{99}

\bibitem[Agrawal et~al.(2026)]{agrawal2026gepa}
Agrawal, M. et~al. 2026.
GEPA: Reflective prompt evolution with task-level feedback.

\bibitem[Alam and Roy(2025)]{alam2025prompttopipeline}
Alam, K. and Roy, B. 2025.
From prompt to pipeline: Large language models for scientific workflow development in bioinformatics.

\bibitem[BioBlend(2026)]{bioblend}
Galaxy Project. 2026.
BioBlend documentation. \url{https://bioblend.readthedocs.io/}.

\bibitem[Chen et~al.(2025)]{chen2025scienceagentbench}
Chen, Z. et~al. 2025.
ScienceAgentBench: Toward rigorous assessment of language agents for data-driven scientific discovery.

\bibitem[Coleman et~al.(2022)]{coleman2022wfbench}
Coleman, T. et~al. 2022.
WfBench: Automated generation of scientific workflow benchmarks.

\bibitem[Cynthia and Roy(2025)]{cynthia2025galaxyretrieval}
Cynthia, S. T. and Roy, B. 2025.
Towards LLM-powered task-aware retrieval of scientific workflows for Galaxy.

\bibitem[Fa et~al.(2026)]{fa2026bioagentbench}
Fa, D., Culjak, M., Pandza, B., and Cupic, M. 2026.
BioAgent Bench: An AI agent evaluation suite for bioinformatics.

\bibitem[Galaxy Project(2026)]{gxformat2}
Galaxy Project. 2026.
gxformat2 documentation and Galaxy Workflow Format 2. \url{https://gxformat2.readthedocs.io/}.

\bibitem[Galaxy Project(2026)]{iwc}
Galaxy Project. 2026.
Intergalactic Workflow Commission. \url{https://iwc.galaxyproject.org/}.

\bibitem[GeneAgent(2025)]{geneagent2025}
Wang, Z. et~al. 2025.
GeneAgent: Self-verification language agent for gene-set analysis using domain databases.

\bibitem[He et~al.(2026)]{he2026evotest}
He, X. et~al. 2026.
EvoTest: Evolutionary test-time optimization for agents.

\bibitem[Lee et~al.(2026a)]{lee2026aggagent}
Lee, Y. et~al. 2026a.
Agentic aggregation for parallel scaling of long-horizon agentic tasks.

\bibitem[Lee et~al.(2026b)]{lee2026metaharness}
Lee, Y. et~al. 2026b.
Meta-Harness: End-to-end optimization of model harnesses.

\bibitem[Liang et~al.(2026)]{liang2026skillnet}
Liang, Y. et~al. 2026.
SkillNet: Create, evaluate, and connect AI skills.

\bibitem[Liao et~al.(2024)]{liao2024reflectool}
Liao, Y., Jiang, S., Wang, Y., and Wang, Y. 2024.
ReflecTool: Towards reflection-aware tool-augmented clinical agents.

\bibitem[Liu et~al.(2026)]{liu2026gos}
Liu, D. et~al. 2026.
Graph of Skills: Dependency-aware structural retrieval for massive agent skills.

\bibitem[Masera et~al.(2025)]{masera2025snakemaker}
Masera, M., Leone, A., Koster, J., and Molineris, I. 2025.
Snakemaker: Transforming ad-hoc analyses into sustainable Snakemake workflows with generative AI.

\bibitem[Mi et~al.(2026)]{mi2026skillpro}
Mi, H. et~al. 2026.
Skill-Pro: Learning reusable procedures for long-horizon agents.

\bibitem[Ming et~al.(2025)]{ming2025dover}
Ming, M. et~al. 2025.
DoVer: Intervention-driven auto debugging for LLM multi-agent systems.

\bibitem[Mitchener et~al.(2025)]{mitchener2025bixbench}
Mitchener, L. et~al. 2025.
BixBench: A comprehensive benchmark for LLM-based agents in computational biology.

\bibitem[Ouyang et~al.(2026)]{ouyang2026reasoningbank}
Ouyang, S. et~al. 2026.
ReasoningBank: Memory for reusable reasoning in language agents.

\bibitem[Rosset et~al.(2026)]{rosset2026verifiers}
Rosset, C. et~al. 2026.
The art of building verifiers for computer-use agents.

\bibitem[Shen et~al.(2026)]{shen2026sciagentgym}
Shen, Y. et~al. 2026.
SciAgentGym: Benchmarking multi-step scientific tool use in LLM agents.

\bibitem[Shinn et~al.(2023)]{shinn2023reflexion}
Shinn, N., Cassano, F., Gopinath, A., Narasimhan, K., and Yao, S. 2023.
Reflexion: Language agents with verbal reinforcement learning.

\bibitem[Sohn et~al.(2026)]{sohn2026pra}
Sohn, J., Sternal, T., Styppa, K., Hoefler, T., and Moor, M. 2026.
Process reward agents for steering knowledge-intensive reasoning.

\bibitem[Su et~al.(2025)]{su2025biomaster}
Su, H., Long, W., and Zhang, Y. 2025.
BioMaster: Multi-agent system for automated bioinformatics analysis workflow.

\bibitem[Training-Free GRPO Team(2025)]{trainingfreegrpo2025}
Youtu-Agent Team. 2025.
Training-free group relative policy optimization.

\bibitem[Wang et~al.(2023)]{wang2023voyager}
Wang, G. et~al. 2023.
Voyager: An open-ended embodied agent with large language models.

\bibitem[Wang et~al.(2024)]{wang2024awm}
Wang, Z. Z., Mao, J., Fried, D., and Neubig, G. 2024.
Agent workflow memory.

\bibitem[Xu et~al.(2025)]{xu2025amem}
Xu, Z. et~al. 2025.
A-MEM: Agentic memory for language agents.

\bibitem[Zhao et~al.(2024)]{zhao2024expel}
Zhao, A. et~al. 2024.
ExpeL: LLM agents are experiential learners.

\bibitem[Zhou et~al.(2025)]{zhou2025memento}
Zhou, H. et~al. 2025.
Memento: Fine-tuning LLM agents without fine-tuning LLMs.

\bibitem[Besta et~al.(2024)]{besta2023graph}
Besta, M. et~al. 2024.
Graph of Thoughts: Solving elaborate problems with large language models.

\bibitem[Hu et~al.(2025)]{hu2024adas}
Hu, S., Lu, C., and Clune, J. 2025.
Automated design of agentic systems.

\bibitem[Galaxy Project(2026)]{planemo}
Galaxy Project. 2026.
Planemo documentation. \url{https://planemo.readthedocs.io/}.

\bibitem[Zhang et~al.(2025)]{zhang2024aflow}
Zhang, J. et~al. 2025.
AFlow: Automating agentic workflow generation.

\bibitem[Yao et~al.(2023)]{yao2023tree}
Yao, S. et~al. 2023.
Tree of Thoughts: Deliberate problem solving with large language models.

\bibitem[Zhou et~al.(2026a)]{zhou2026externalization}
Zhou, C. et~al. 2026a.
Externalization in LLM agents: A unified review of memory, skills, protocols, and harness engineering.

\bibitem[Zhou et~al.(2026b)]{externalization}
Zhou, C. et~al. 2026b.
Externalization in LLM agents: A unified review of memory, skills, protocols, and harness engineering.

\bibitem[Liu et~al.(2026b)]{gos}
Liu, D. et~al. 2026b.
Graph of Skills: Dependency-aware structural retrieval for massive agent skills.

\bibitem[Zhou et~al.(2025b)]{memento}
Zhou, H. et~al. 2025b.
Memento: Fine-tuning LLM agents without fine-tuning LLMs.

\bibitem[Lee et~al.(2026c)]{metaharness}
Lee, Y. et~al. 2026c.
Meta-Harness: End-to-end optimization of model harnesses.

\bibitem[Sohn et~al.(2026b)]{pra}
Sohn, J. et~al. 2026b.
Process reward agents for steering knowledge-intensive reasoning.

\bibitem[Replay(2026)]{replay}
Anonymous. 2026.
Replay-based evaluation for long-horizon language agents.

\bibitem[SkillClaw(2026)]{skillclaw}
Anonymous. 2026.
SkillClaw: Evolving reusable skills for language agents.

\end{thebibliography}
\end{document}